\documentclass[twoside,leqno,twocolumn]{article}

\usepackage[letterpaper]{geometry}

\usepackage{ltexpprt}
\usepackage{hyperref}

\usepackage[utf8]{inputenc} 
\usepackage[T1]{fontenc}    
\usepackage{hyperref}       
\usepackage{url}            
\usepackage{booktabs}       
\usepackage{amsfonts}       
\usepackage{nicefrac}       
\usepackage{microtype}      
\usepackage[table,dvipsnames]{xcolor}        

\usepackage{epsfig}
\usepackage{wrapfig}

\usepackage{times}

\usepackage[small]{caption}
\usepackage{graphicx}
\usepackage{amsmath}
\usepackage{booktabs}
\usepackage{multirow}
\usepackage{stmaryrd}

\usepackage[normalem]{ulem}
\useunder{\uline}{\ul}{}

\usepackage[utf8]{inputenc}
\usepackage{pgfplots}
\DeclareUnicodeCharacter{2212}{−}
\usepgfplotslibrary{groupplots,dateplot}
\usetikzlibrary{patterns,shapes.arrows}
\pgfplotsset{compat=newest}

\title{Continuously Generalized Ordinal Regression for Linear and Deep Models}

\author{
  Fred Lu\thanks{Booz Allen Hamilton. Emails: {lu\_fred,raff\_edward@bah.com}}\ \thanks{Univ. of Maryland, Baltimore County}
  \and 
  Francis Ferraro\footnotemark[2]
  \and
  Edward Raff\footnotemark[1]\ \footnotemark[2]
}

\date{}

\usepackage{array}
\usepackage{amsmath}
\usepackage{mathtools}
\usepackage{amsfonts} 
\usepackage{graphicx}
\usepackage{todonotes}
\usepackage{tikz}
\usepackage{enumitem}
\usepackage{placeins}
\usetikzlibrary{arrows}
\usepackage{url}
\usepackage{adjustbox}

\usepackage{physics}
\usepackage{mathdots}
\usepackage{yhmath}
\usepackage{cancel}
\usepackage{color}
\usepackage{siunitx}
\usepackage{amssymb}
\usepackage{gensymb}
\usepackage{tabularx}
\usetikzlibrary{fadings}
\usetikzlibrary{patterns}
\usetikzlibrary{shadows.blur}
\usetikzlibrary{shapes}
\usepackage[outline]{contour}

\usepackage{fancyhdr}

\everypar{\looseness=-1} 
\begin{document}

\maketitle

\fancyfoot[R]{\scriptsize{Copyright \textcopyright\ 2022 by SIAM\\
Unauthorized reproduction of this article is prohibited}}

\begin{abstract} \small\baselineskip=9pt
Ordinal regression is a classification task where classes have an order and prediction error increases the further the predicted class is from the true class.
The standard approach for modeling ordinal data involves fitting parallel separating hyperplanes that optimize a certain loss function.
This assumption offers sample efficient learning via inductive bias, but is often too restrictive in real-world datasets where features may have varying effects across different categories.
Allowing class-specific hyperplane slopes creates generalized logistic ordinal regression, increasing the flexibility of the model at a  cost to sample efficiency.
We explore an extension of the generalized model to the all-thresholds logistic loss and propose a regularization approach that interpolates between these two extremes.
Our method, which we term \textit{continuously generalized ordinal logistic}, significantly outperforms the standard ordinal logistic model over a thorough set of ordinal regression benchmark datasets.
We further extend this method to deep learning and show that it achieves competitive or lower prediction error compared to previous models over a range of datasets and modalities. Furthermore, two primary alternative models for deep learning ordinal regression are shown to be special cases of our framework.

\end{abstract}

\section{Introduction}

In ordinal regression problems, the prediction task is to choose the target $y$ from a set of labels with an ordered relation, e.g. $1 \prec 2 \prec \ldots \prec k$. Unlike in classification, where accuracy is paramount, in ordinal regression the loss generally increases as the model predicts classes further away from the true label. 
Consider predicting medication dosage, where adjacent dosage amounts may still be safe, but large differences in dosage can be fatal.
Ordinal regression models may give more accurate answers and learn more efficiently than classifiers in these situations because their inductive bias is better suited to the problem. 

\begin{figure}[!ht]
    \centering
    \adjustbox{max width=0.99\columnwidth}{%
    \input{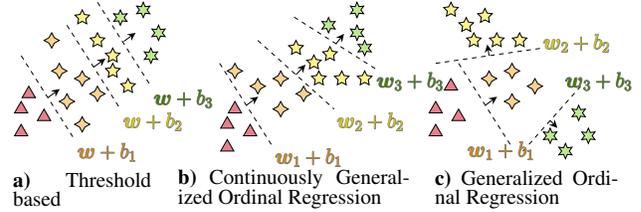}
    }
    \caption{Figure (a) shows the standard threshold based approach on left which assumes all hyper-planes are equal, and (c) on the right shows the generalized ordinal regression which makes no assumption on the relationship between hyper-planes. Our approach coGOL allows for a continuous relaxation on the relationship between hyper-planes, allowing it to model cases that have variable relationship between hyper-planes, and captures both (a) and (c) as special cases of infinite and zero regularization respectively. }
    \label{fig:ordinal_diagram}
\end{figure}

Threshold-based models are one of the most widely used ordinal regression approaches. In such models, a continuous prediction is learned as a linear mapping from the features along with a set of thresholds which partitions the prediction into classes \cite{mccullagh1980regression, chu2007support}. While such models are simple and efficient to learn, the restriction often imposes unreasonably strict requirements on the nature of the ordinal relationship of the data. To address this restriction, some ordinal models introduce generalized coefficients, learning a separate linear mapping for each class \cite{williams2016understanding,ananth1997regression}. Such models can be viewed as training a separate binary classifier for each cut-point in the class ordering \cite{gutierrez2015ordinal}. While this adds much-needed flexibility, it can exhibit poor stability on many datasets especially in high-noise regimes. Models of this type must also combine separate binary predictions into a consistent ordinal prediction~\cite{frank2001simple}.

To address these issues, we propose a continuous interpolation between the threshold and the generalized ordinal models, called \textit{continuously generalized ordinal logit} (coGOL). By introducing generalized coefficients to the \textit{all-thresholds} ordinal loss function with a new regularization approach, \mbox{coGOL} combines the flexibility and stability of the two extreme prior approaches as shown in \autoref{fig:ordinal_diagram}.
By being able to adjust model flexibility, coGOL obtains better linear performance on 17 established benchmark datasets, with competitive results in deep models as well. Our extensive experiments show that coGOL achieves a statistically significant improvement over the standard ordinal logistic benchmark. This makes coGOL an ordinal approach appropriate for a wider variety of scenarios than prior methods. 

Our method connects previous frameworks for threshold-based ordinal regression. We provide an overview and discuss previous approaches in \autoref{sec:related_work}. The formulation of our method, with linear, kernelized, and deep versions, is given in \autoref{sec:cogol}, and we detail our experimental evaluations in \autoref{sec:experiments}. In \autoref{sec:results} we find that linear coGOL significantly outperforms the standard ordinal logistic baseline over 17 benchmark datasets, while deep coGOL is competitive with or better than other deep learning approaches on a set of large-scale, image-based, and sequence-based data. We also implement kernelized versions of our models and show that linear coGOL has advantages even compared to the kernel setting. Finally we conclude in \autoref{sec:conclusion}.

\section{Related Work} \label{sec:related_work}

\subsection{Ordinal regression methods}

Traditional ordinal regression approaches can be grouped into three categories as suggested by \cite{gutierrez2015ordinal}. In the first category, one simply fits a classification model, or a regression model whose output is then discretized. While straight-forward, this basic approach does not properly account for the ordinal regression problem. Since ordinal regression falls somewhere between classification and regression, the problem benefits from more specialized approaches. Cost-sensitive classification is a multi-class extension that specifies a cost matrix with larger loss as classes are further apart \cite{lin2008ordinal}. While the cost structure is ostensibly similar to the models we study, the underlying model is based on softmax classification and does not leverage the ordinal association between feature and target \cite{lin2012reduction}.

A second category decomposes the ordinal categories into binary classification problems, whose outputs can be combined into an ordinal prediction. Approaches to modeling and combining the outputs may involve support vector machines (SVM)~\cite{frank2001simple, waegeman2009ensemble, wang2017nonparallel} or neural networks~\cite{cheng2008neural,niu2016ordinal}.

The final category contains threshold-based approaches, which simultaneously learn an output mapping and appropriate thresholds that partition the output to make ordinal predictions. For example, linear threshold models generally find the parallel hyperplanes that best separate the ordinal classes by minimizing an objective function, which include variants of SVM hinge loss and logistic loss \cite{chu2007support,rennie2005loss,chu2005gaussian,lin2012reduction}. 

A well-known class of linear threshold models, known as \textit{cumulative logit} or \textit{proportional odds}, comes from the statistics literature and is imbued with probabilistic interpretation as a latent continuous target variable under censoring \cite{mccullagh1980regression,agresti2013modeling}. Mathematically, the ordinal target $y$ is modeled as
$\mathbb{P}(y \leq j | \boldsymbol{x}) = \frac{\exp(\theta_j - \boldsymbol{w}^\top  \boldsymbol{x})}{1 + \exp(\theta_j - \boldsymbol{w}^\top \boldsymbol{x})}$. 
Among many variants of this model, one generalizes the weights to be class-dependent, replacing $\boldsymbol{w}$ above with $\boldsymbol{w}_j$ \cite{ananth1997regression}, and is known as the \textit{generalized ordered logit} model \cite{williams2016understanding}. While this is an important starting point for our work, the underlying proportional odds assumption of this class of models does not generalize well to many datasets.

A unified framework for linear threshold losses is developed in \cite{rennie2005loss,pedregosa2017consistency} and interprets the objective function as a convex surrogate for a natural loss function. Two such losses are the mean absolute error and the 0-1 loss and their corresponding surrogates were named the \textit{all-thresholds} and \textit{immediate-threshold} loss. Intuitively, the difference between the two versions is that the all-thresholds loss on an $(x, y)$ pair contains penalty terms from all ordinal classes, regardless of the value of $y$, while the immediate-threshold loss only contains terms from the thresholds corresponding to $y$. As the mean absolute error better fits the principle that the larger the misprediction, the more \textit{wrong} the model is, this lends theoretical support for using the all-thresholds loss. This is supported by experimental evidence in \cite{rennie2005loss}. Interestingly, the log-likelihood structure of the cumulative logit model is similar to the \textit{immediate-thresholds} loss \cite{pedregosa2017consistency}. We clarify these details in our Appendix.

Our work first introduces generalized coefficients to the threshold model using the all-thresholds loss. In preliminary experiments, we found that the generalized model using immediate-threshold loss had unstable results, specifically with \textit{higher} training and validation error than the standard ordinal logit, despite being a larger model class. While this step alone improves results on many datasets, we also add a regularization approach that permits control over the flexibility of the model. This is a novel addition which allows a bias-variance tradeoff in the model, thus combatting under- and over-fitting. Our study interpolates between the standard and generalized ordinal models using the all-thresholds loss in a novel manner that addresses shortcomings of the originals, by rewarding models which smoothly vary their hyperplane directions. This matches an inductive prior for many real-world datasets where features relate monotonically with the ordinal target, but vary to a greater extent over certain subsets of ordinal classes.

Prior works demonstrate a reduction from ordinal all-threshold models to binary classification with a specific class cost matrix \cite{li2006ordinal, lin2012reduction}. In fact, a generalized coefficient threshold model would reduce to classification with separate weights for each ordinal class divide. Thus our approach forms a bridge between the second (binary decomposition) and third (threshold) approach, combining flexibility and stability.



\subsection{Deep ordinal regression}

Deep learning approaches generally also fall into the three previously described categories. Some methods handle ordinal regression as a classification problem \cite{levi2015age,rothe2015dex}, while an early proposed neural network for ordinal regression adopted a threshold approach \cite{mathieson1996ordinal}. Similar to the cost-sensitive approaches in the linear case, deep models have found success with using weighted softmax labels to formulate a classification problem \cite{diaz2019soft}.

Following the binary decomposition approach of \cite{li2006ordinal,cheng2008neural}, a deep learning model with multiple outputs was developed by \cite{niu2016ordinal} and showed strong performance on face age estimation datasets. The model (named OR-CNN) shares weights through all intermediate layers and ends with separate binary classification layers for each threshold $\theta_j$, where each layer predicts whether the sample's rank is higher than $\theta_j$.

This framework was modified in \cite{cao2019rank} with additional weight sharing in the classification layers, so that only the biases differ. They then proved their model (named CORAL) to be order-consistent and to improve performance on age-estimation datasets. This model falls in the threshold model category.

In our work, we attach the continuously generalized ordinal model with the all-thresholds loss at the end of a neural network. Our model structure is in fact similar to OR-CNN, but our model further permits interpolation between the generalized and standard ordinal model. In fact, we will later show that both OR-CNN and CORAL are special cases of our framework. OR-CNN is equivalent to our generalized ordinal logit model with unconstrained weights. In contrast, CORAL is equivalent to the \textit{all-thresholds} loss with non-generalized (parallel) thresholds. We will note that OR-CNN outperforms CORAL on all datasets, and performs more competitively with our coGOL model.

Another approach for deep ordinal regression \cite{liu2018constrained} implements ordinal regression as a weighted combination of an unconstrained logistic regression and a monotonic threshold constraint based on the Hinge loss, which requires balancing the mismatched gradient scales between the two different loss functions. Where available, we include their results as a baseline.

A recent concurrent work in deep ordinal learning has also proposed regularization on the flexibility of generalized coefficients \cite{guo2021order}. However, their regularization terms are perhaps overly complex, depending on pairwise cosine similarities between weight vectors, variance of weight norms, and constraints on bias terms. In order to balance these objectives, at least 5 separate hyperparameters are required. In contrast, our method requires one new hyperparameter and provides a much more intuitive generalization of the ordinal model. We implement their method as a baseline.

\section{Continuously Generalized Ordinal Logistic} \label{sec:cogol}

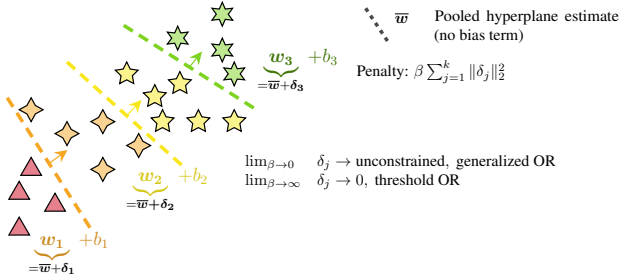
\begin{figure}[!ht]
    \centering
    \vspace{-15pt}
    \adjustbox{max width=0.99\columnwidth}{%
    \tikzset{every picture/.style={line width=0.75pt}} 

\begin{tikzpicture}[x=0.75pt,y=0.75pt,yscale=-1,xscale=1]

\draw  [fill={rgb, 255:red, 245; green, 166; blue, 35 }  ,fill opacity=0.5 ] (70,105) -- (73.54,111.46) -- (80,115) -- (73.54,118.54) -- (70,125) -- (66.46,118.54) -- (60,115) -- (66.46,111.46) -- cycle ;
\draw  [fill={rgb, 255:red, 248; green, 231; blue, 28 }  ,fill opacity=0.5 ] (120,55) -- (122.94,60.95) -- (129.51,61.91) -- (124.76,66.55) -- (125.88,73.09) -- (120,70) -- (114.12,73.09) -- (115.24,66.55) -- (110.49,61.91) -- (117.06,60.95) -- cycle ;
\draw  [fill={rgb, 255:red, 126; green, 211; blue, 33 }  ,fill opacity=0.5 ] (162.68,25) -- (165.18,30.67) -- (171.34,30) -- (167.68,35) -- (171.34,40) -- (165.18,39.33) -- (162.68,45) -- (160.18,39.33) -- (154.02,40) -- (157.68,35) -- (154.02,30) -- (160.18,30.67) -- cycle ;
\draw  [fill={rgb, 255:red, 245; green, 166; blue, 35 }  ,fill opacity=0.5 ] (130,115) -- (133.54,121.46) -- (140,125) -- (133.54,128.54) -- (130,135) -- (126.46,128.54) -- (120,125) -- (126.46,121.46) -- cycle ;
\draw  [fill={rgb, 255:red, 245; green, 166; blue, 35 }  ,fill opacity=0.5 ] (100,95) -- (103.54,101.46) -- (110,105) -- (103.54,108.54) -- (100,115) -- (96.46,108.54) -- (90,105) -- (96.46,101.46) -- cycle ;
\draw  [fill={rgb, 255:red, 248; green, 231; blue, 28 }  ,fill opacity=0.5 ] (143.5,74) -- (146.44,79.95) -- (153.01,80.91) -- (148.26,85.55) -- (149.38,92.09) -- (143.5,89) -- (137.62,92.09) -- (138.74,85.55) -- (133.99,80.91) -- (140.56,79.95) -- cycle ;
\draw  [fill={rgb, 255:red, 248; green, 231; blue, 28 }  ,fill opacity=0.5 ] (150,95) -- (152.94,100.95) -- (159.51,101.91) -- (154.76,106.55) -- (155.88,113.09) -- (150,110) -- (144.12,113.09) -- (145.24,106.55) -- (140.49,101.91) -- (147.06,100.95) -- cycle ;
\draw  [fill={rgb, 255:red, 126; green, 211; blue, 33 }  ,fill opacity=0.5 ] (200,5) -- (202.5,10.67) -- (208.66,10) -- (205,15) -- (208.66,20) -- (202.5,19.33) -- (200,25) -- (197.5,19.33) -- (191.34,20) -- (195,15) -- (191.34,10) -- (197.5,10.67) -- cycle ;
\draw  [fill={rgb, 255:red, 126; green, 211; blue, 33 }  ,fill opacity=0.5 ] (210,55) -- (212.5,60.67) -- (218.66,60) -- (215,65) -- (218.66,70) -- (212.5,69.33) -- (210,75) -- (207.5,69.33) -- (201.34,70) -- (205,65) -- (201.34,60) -- (207.5,60.67) -- cycle ;
\draw [color={rgb, 255:red, 245; green, 166; blue, 35 }  ,draw opacity=1 ][line width=2.25]  [dash pattern={on 6.75pt off 4.5pt}]  (20,85) -- (90,195) ;
\draw [color={rgb, 255:red, 248; green, 231; blue, 28 }  ,draw opacity=1 ][line width=2.25]  [dash pattern={on 6.75pt off 4.5pt}]  (70,55) -- (170,145) ;
\draw  [fill={rgb, 255:red, 245; green, 166; blue, 35 }  ,fill opacity=0.5 ] (100,135) -- (103.54,141.46) -- (110,145) -- (103.54,148.54) -- (100,155) -- (96.46,148.54) -- (90,145) -- (96.46,141.46) -- cycle ;
\draw  [fill={rgb, 255:red, 248; green, 231; blue, 28 }  ,fill opacity=0.5 ] (164.51,63.09) -- (167.45,69.05) -- (174.02,70) -- (169.27,74.64) -- (170.39,81.18) -- (164.51,78.09) -- (158.63,81.18) -- (159.76,74.64) -- (155,70) -- (161.57,69.05) -- cycle ;
\draw  [fill={rgb, 255:red, 248; green, 231; blue, 28 }  ,fill opacity=0.5 ] (180,95) -- (182.94,100.95) -- (189.51,101.91) -- (184.76,106.55) -- (185.88,113.09) -- (180,110) -- (174.12,113.09) -- (175.24,106.55) -- (170.49,101.91) -- (177.06,100.95) -- cycle ;
\draw [color={rgb, 255:red, 126; green, 211; blue, 33 }  ,draw opacity=1 ][line width=2.25]  [dash pattern={on 6.75pt off 4.5pt}]  (117.5,24.33) -- (227.5,94.33) ;
\draw  [fill={rgb, 255:red, 248; green, 231; blue, 28 }  ,fill opacity=0.5 ] (210,95) -- (212.94,100.95) -- (219.51,101.91) -- (214.76,106.55) -- (215.88,113.09) -- (210,110) -- (204.12,113.09) -- (205.24,106.55) -- (200.49,101.91) -- (207.06,100.95) -- cycle ;
\draw  [fill={rgb, 255:red, 126; green, 211; blue, 33 }  ,fill opacity=0.5 ] (200,35) -- (202.5,40.67) -- (208.66,40) -- (205,45) -- (208.66,50) -- (202.5,49.33) -- (200,55) -- (197.5,49.33) -- (191.34,50) -- (195,45) -- (191.34,40) -- (197.5,40.67) -- cycle ;
\draw [color={rgb, 255:red, 245; green, 166; blue, 35 }  ,draw opacity=1 ]   (55,140) -- (68.07,130.51) ;
\draw [shift={(70.5,128.75)}, rotate = 504.03] [fill={rgb, 255:red, 245; green, 166; blue, 35 }  ,fill opacity=1 ][line width=0.08]  [draw opacity=0] (10.72,-5.15) -- (0,0) -- (10.72,5.15) -- (7.12,0) -- cycle    ;
\draw [color={rgb, 255:red, 248; green, 231; blue, 28 }  ,draw opacity=1 ][fill={rgb, 255:red, 248; green, 231; blue, 28 }  ,fill opacity=1 ]   (120,100) -- (130.44,88.93) ;
\draw [shift={(132.5,86.75)}, rotate = 493.33] [fill={rgb, 255:red, 248; green, 231; blue, 28 }  ,fill opacity=1 ][line width=0.08]  [draw opacity=0] (10.72,-5.15) -- (0,0) -- (10.72,5.15) -- (7.12,0) -- cycle    ;
\draw [color={rgb, 255:red, 126; green, 211; blue, 33 }  ,draw opacity=1 ]   (172.5,59.33) -- (181.12,48.59) ;
\draw [shift={(183,46.25)}, rotate = 488.76] [fill={rgb, 255:red, 126; green, 211; blue, 33 }  ,fill opacity=1 ][line width=0.08]  [draw opacity=0] (10.72,-5.15) -- (0,0) -- (10.72,5.15) -- (7.12,0) -- cycle    ;
\draw  [fill={rgb, 255:red, 208; green, 2; blue, 27 }  ,fill opacity=0.5 ] (30,155) -- (34.33,162.5) -- (38.66,170) -- (30,170) -- (21.34,170) -- (25.67,162.5) -- cycle ;
\draw  [fill={rgb, 255:red, 208; green, 2; blue, 27 }  ,fill opacity=0.5 ] (30,185) -- (34.33,192.5) -- (38.66,200) -- (30,200) -- (21.34,200) -- (25.67,192.5) -- cycle ;
\draw  [fill={rgb, 255:red, 208; green, 2; blue, 27 }  ,fill opacity=0.5 ] (60,165) -- (64.33,172.5) -- (68.66,180) -- (60,180) -- (51.34,180) -- (55.67,172.5) -- cycle ;
\draw  [fill={rgb, 255:red, 208; green, 2; blue, 27 }  ,fill opacity=0.5 ] (40,135) -- (44.33,142.5) -- (48.66,150) -- (40,150) -- (31.34,150) -- (35.67,142.5) -- cycle ;
\draw  [fill={rgb, 255:red, 245; green, 166; blue, 35 }  ,fill opacity=0.5 ] (40,85) -- (43.54,91.46) -- (50,95) -- (43.54,98.54) -- (40,105) -- (36.46,98.54) -- (30,95) -- (36.46,91.46) -- cycle ;
\draw [color={rgb, 255:red, 74; green, 74; blue, 74 }  ,draw opacity=1 ][line width=2.25]  [dash pattern={on 2.53pt off 3.02pt}]  (320,10) -- (340,40) ;

\draw (36.33,195.9) node [anchor=north west][inner sep=0.75pt]  [font=\large,color={rgb, 255:red, 203; green, 137; blue, 28 }  ,opacity=1 ]  {$\underbrace{\boldsymbol{w_{1}}}_{\textcolor[rgb]{0,0,0}{=\overline{\boldsymbol{w}} +\boldsymbol{\delta _{1}}}} +b_{1}$};
\draw (118,142.4) node [anchor=north west][inner sep=0.75pt]  [font=\large,color={rgb, 255:red, 213; green, 199; blue, 23 }  ,opacity=1 ]  {$\underbrace{\boldsymbol{w_{2}}}_{\textcolor[rgb]{0,0,0}{=\boldsymbol{\overline{\boldsymbol{w}} +\delta _{2}}}} +b_{2}$};
\draw (228,42.4) node [anchor=north west][inner sep=0.75pt]  [font=\large,color={rgb, 255:red, 65; green, 117; blue, 5 }  ,opacity=1 ]  {$\underbrace{\boldsymbol{w_{3}}}_{\textcolor[rgb]{0,0,0}{=\boldsymbol{\overline{w}} +\boldsymbol{\delta _{3}}}} +b_{3}$};
\draw (343,12.4) node [anchor=north west][inner sep=0.75pt]  [color={rgb, 255:red, 0; green, 0; blue, 0 }  ,opacity=1 ]  {$\boldsymbol{\overline{w}}$};
\draw (455,25) node   [align=left] {\begin{minipage}[lt]{115.60000000000001pt}\setlength\topsep{0pt}
Pooled hyperplane estimate \ (no bias term)
\end{minipage}};
\draw (311,52.4) node [anchor=north west][inner sep=0.75pt]    {$\text{Penalty: } \beta \sum ^{k}_{j=1}\Vert \delta _{j}\Vert ^{2}_{2}$};
\draw (211,129.4) node [anchor=north west][inner sep=0.75pt]    {$\begin{array}{ l l }
\lim _{\beta \rightarrow 0} & \delta _{j}\rightarrow \text{unconstrained} ,\ \text{generalized OR}\\
\lim _{\beta \rightarrow \infty } & \delta _{j}\rightarrow 0,\ \text{threshold OR}
\end{array}$};

\end{tikzpicture}
    }
    \caption{
    The essence of our approach---coGOL---is to represent the solution as a single ``pooled'' weight vector $\bar{\boldsymbol{w}}$
    with each class's weight vector $\boldsymbol{w}_j$ being an adjustment
    $\boldsymbol{w}_j = \bar{\boldsymbol{w}} + \boldsymbol{\delta}_j$ for each class. 
    A penalty $\beta$ on $\|\boldsymbol{\delta}_j\|$ controls how much each class's solution may deviate from the pooled estimate $\bar{\boldsymbol{w}}$, and an unregularized sequence of biases shift the class hyper-planes.
    }
    \label{fig:cogol}
\end{figure}

Suppose we have features $\boldsymbol{x} \in \mathbb{R}^p$ and an ordered target variable $y \in \{1,2,\ldots,k\}$ as defined previously. Our goal is to learn a decision function $f:\mathbb{R}^{p} \to \mathbb{R}$ that minimizes the risk $\mathcal{L}(f) = \mathbb{E}[\ell(y, f(\boldsymbol{x}))]$ under some loss function $\ell$.

Linear ordinal regression models restrict $f$ to the set of parallel linear models $\{g(\boldsymbol{x}; \boldsymbol{w}, \theta)\}$ where $g_j(\boldsymbol{x}) = \theta_j - \boldsymbol{w}^\top \boldsymbol{x}$ with bias terms $\theta \in \mathbb{R}^{k-1}$ such that $\theta_1 \leq \ldots \leq \theta_{k-1}$. That is, the model is a set of $k-1$ thresholds that partition the linear output $\boldsymbol{w}^\top \boldsymbol{x}$ into the ordinal classes. Prediction is then determined by the number of thresholds crossed:
\begin{equation} \label{eq1}
    f(\boldsymbol{x}) = 1 + \sum_{j=1}^{k-1} \llbracket g_j(\boldsymbol{x}) < 0 \rrbracket
\end{equation}
\noindent
where the brackets $\llbracket \cdot \rrbracket$ denote the indicator function.
A natural choice for the loss is mean absolute error: $\ell(y, x) = |y - x|$. Under the linear model class, the loss can be rewritten as
$$\ell(y, f(\boldsymbol{x})) = \sum_{j=1}^{y-1} \llbracket g_j(\boldsymbol{x}) \geq 0 \rrbracket + \sum_{j=y}^{k-1} \llbracket g_j(\boldsymbol{x}) < 0 \rrbracket$$
as shown in \cite{pedregosa2017consistency}. Next, we step through two core aspects of coGOL: a continuous loss and generalized coefficients.

\paragraph{A Continuous Loss} Since the loss is discontinuous, we replace the indicator functions with a convex, continuous surrogate for optimization. In this work we focus on the logistic loss: $\varphi(x) = \log(1 + e^{-x})$. The loss then becomes
\begin{equation} \label{eq2}
    \ell(y, f(\boldsymbol{x})) = \sum_{j=1}^{y-1} \varphi(-g_j(\boldsymbol{x})) + \sum_{j=y}^{k-1} \varphi(g_j(\boldsymbol{x}))
\end{equation}
also known as the \textit{all-thresholds} loss. Optimization of this loss function under monotonic $\theta$ results in the standard all-thresholds \textit{ordinal logit} (OL) model. The monotonicity is enforced by formulating it as a convex optimization problem with the constraint on $\theta$.

\paragraph{Generalized Coefficients}
We introduce generalized coefficients by extending the model class to $\tilde{g}_j(\boldsymbol{x}) = \theta_j - \boldsymbol{w}_j^\top \boldsymbol{x}$, so that the threshold hyperplanes need not be parallel anymore. This yields the all-thresholds version of the \textit{generalized ordinal logit} (GOL) model.

Finally, we wish to continuously interpolate between the parallel and generalized models in a manner that rewards smoothness between neighboring classes. Allowing $\boldsymbol{\delta}_j = \boldsymbol{w}_j - \boldsymbol{w}_{j-1}$, we penalize the magnitude of $\boldsymbol{\delta}_j$.
As in standard OL models, we also permit L2 regularization on $\boldsymbol{w_j}$. Altogether, this results in the \textit{continously generalized ordinal logit} (coGOL) loss function:
\begin{equation} \label{eq:cogol}
 \begin{split}
     \ell(y, f(\boldsymbol{x})) =& \sum_{j=1}^{y-1} \varphi(- \tilde{g}_j(\boldsymbol{x}) ) + \sum_{j=y}^{k-1} \varphi( \tilde{g}_j(\boldsymbol{x})) \\
     &+ \underbrace{\alpha \sum_{j=1}^{k-1} \|\boldsymbol{w}_j\|_2^2}_{\text{standard L2 penalty} } + 
     \underbrace{\beta \sum_{j=2}^{k-1}\|\boldsymbol{\delta}_j\|_2^2}_{\text{our deviation penalty}}.
 \end{split}
\end{equation}
As the penalty $\beta \rightarrow \infty$, we force $\boldsymbol{\delta}_j \rightarrow 0$, recovering the threshold model. At $\beta = 0$ we recover the GOL model, and all $\beta$ in between give us a continuous spectrum interpolating between these two models. We construct toy datasets to demonstrate how \autoref{eq:cogol} can make significant alterations to the learned solution, capturing the sample efficiency of OL and the expressiveness of GOL (Appendix 1).







\subsection{Deep learning formulation}

We extend the continuously generalized all-thresholds loss to deep learning by replacing $\boldsymbol{x}$ with the output of a base neural network $F(\boldsymbol{x})$. We can view $F(\boldsymbol{x})$ as a feature extractor trained alongside the ordinal regression head. We implement the ordinal regression as a final linear layer with $k-1$ independent weight vectors and $k-1$ biases and train the model using the loss of \autoref{eq:cogol}. For convenience, we re-express the loss using familiar deep learning notation, with $\sigma$ as the sigmoid function:
\begin{equation} \label{eq4}
    \begin{split}
        \ell(&y, f(\boldsymbol{x})) = - \sum_{j=1}^{y-1} \log\sigma( - \tilde{g}_j(F(\boldsymbol{x})) )\\
            &- \sum_{j=y}^{k-1} \log\sigma( \tilde{g}_j(F(\boldsymbol{x})) )  + \alpha \sum_{j=1}^{k-1} \|\boldsymbol{w}_j\|_2^2 + \beta \sum_{j=2}^{k-1}\|\boldsymbol{\delta}_j\|_2^2.
    \end{split}
\end{equation}

We note that setting $\alpha>0$ has an important additional effect of constraining the magnitude of the weights $\boldsymbol{w}_j$ besides standard weight decay. Otherwise for any value of $\beta>0$, the feature extractor $F(\boldsymbol{x})$ can rescale the output to counteract the constraint on $\boldsymbol{\delta}_j$.

The linear coGOL was implemented with \textit{cvxpy} in Python 3.6 \cite{cvxpy,agrawal2018rewriting}, while the deep learning models were implemented in Pytorch 1.6. We benchmarked our linear OL model with the \textit{mord} package to ensure they gave identical results \cite{pedregosa2017consistency}.

\subsection{Connection to OR-CNN and CORAL}
While both OR-CNN \cite{niu2016ordinal} and CORAL \cite{cao2019rank} were developed as CNNs for age estimation, we adapt their models and losses for general neural networks as comparison methods. In this section we rewrite their loss functions to demonstrate their equivalence to the extreme cases of the coGOL model. Both works borrow the convention of \cite{li2006ordinal}, converting the ordinal label $y$ into an extended binary label $\tilde{y}\in \mathbb{R}^{k-1}$, such that $\tilde{y}_j = \llbracket y > j \rrbracket$. 

In CORAL, the output of the base network can be written as $h_j(\boldsymbol{x}) = \theta_j +\boldsymbol{w} \cdot F(\boldsymbol{x})$. Under uniform sample weights, their loss function $\ell(y, f(\boldsymbol{x}))$ is then
\begin{equation*}
\begin{aligned}
&=-\sum ^{k-1}_{j=1}\log \sigma (h_{j} (\boldsymbol{x} )) \tilde{y}_{j}  +\log (1-\sigma (h_{j} (\boldsymbol{x} ))(1-\tilde{y}_{j} ))\\
&=-\sum ^{k-1}_{j=1}\log \sigma (h_{j} (\boldsymbol{x} ))\tilde{y}_{j}  + \log (\sigma (-h_{j} (\boldsymbol{x} ))(1-\tilde{y}_{j} ))\\
&=-\sum ^{y-1}_{j=1}\log (\sigma (h_{j} (\boldsymbol{x} )))-\sum ^{k-1}_{j=y}\log (\sigma (-h_{j} (\boldsymbol{x} )))
\end{aligned}
\end{equation*}
\noindent
which is equivalent to the all-thresholds loss by setting $g_j(\boldsymbol{x}) = -h_j(\boldsymbol{x})$. In particular, because the coefficient $\boldsymbol{w}$ is not generalized, this is the standard OL model (or coGOL with $\beta \to \infty)$.

In OR-CNN, the penultimate layer connects to $K-1$ separate binary cross-entropy outputs. Each such output $o_j$ is thus the dot product of a weight vector with the penultimate layer outputs plus bias, so equivalently $\tilde{h}_{j}(\boldsymbol{x}) = \theta_j + \boldsymbol{w}_j \cdot F(\boldsymbol{x})$. Under uniform sample and task weights, their loss $\ell(y, f(\boldsymbol{x}))$ is then 
\begin{equation*}
\begin{aligned}
& \begin{split}= -\sum_{j=1}^{k-1} \bigg(    
&\llbracket o_j = \tilde{y}_j \rrbracket \log p(o_j|\boldsymbol{x}, F) + \\ &(1 - \llbracket o_j = \tilde{y}_j \rrbracket) \log(1 - p(o_j| \boldsymbol{x}, F) )
\bigg) \end{split}\\
&=  -\sum_{j=1}^{y-1} \log p(o_j|\boldsymbol{x}, F) - \sum_{j=y}^{k-1} \log(1 - p(o_j | \boldsymbol{x}, F))\\
&= -\sum_{j=1}^{y-1} \log \sigma( \tilde{h}_j(\boldsymbol{x})) - \sum_{j=y}^{k-1} \log(\sigma(- \tilde{h}_j(\boldsymbol{x}))).
\end{aligned}
\end{equation*}
Setting $\tilde{g}_j(\boldsymbol{x}) = -\tilde{h}_j(\boldsymbol{x})$, we get the generalized coefficient loss of \autoref{eq4} but without our deviation penalty. Therefore, OR-CNN is equivalent to coGOL with $\beta=0$.

\section{Experiments} \label{sec:experiments}

\subsection{Linear model benchmarks} \label{sec:linear_experiments}

We evaluate our method using a standard set of 17 ordinal 
datasets as defined in \cite{gutierrez2015ordinal}. Characteristics of each dataset 
are 
in  \autoref{tab:info}.  We replicate our models using the same 30 train-test splits. We use stratified 3-fold cross-validation within each training set to tune regularization parameters $\beta$ and $\alpha$ using 30 iterations of Optuna \cite{akiba2019optuna}. $\alpha$ and $\beta$ were searched in the range [1e-6, 10]. 

We compare the MAE, MSE, and accuracy, averaged over the 30 replications, of coGOL with OL, using the Wilcoxon signed rank test \cite{Wilcoxon1945} to statistically compare the models. Multiple previous studies on evaluation and model comparison have demonstrated the Wilcoxon test over multiple datasets to be more reliable for establishing improvement ~\cite{JMLR:v17:benavoli16a,Demsar:2006:SCC:1248547.1248548}, in comparison to standard error/confidence intervals in repeated trials~\cite{Varma2006,Bengio:2004:NUE:1005332.1044695,Blum:1999:BHB:307400.307439,Alpaydin:1999:CTC:339993.339999,Dietterich:1998:AST:303222.303237}.
Due to the large number of datasets and large number of trials to ensure a statistically valid conclusion, we limit our experiments to logistic loss with $\varphi(x) = \log(1+e^{-x})$. Our framework is fully compatible with other choices such as the Hinge loss (i.e., SVM), but comparing different link functions $\varphi(\cdot)$ is beyond the scope of our work. Practitioners may choose the link function $\varphi(\cdot)$ based on what is  appropriate for their problem. 

\subsection{Kernel model benchmarks}
\begin{table}
\centering
\caption{Characteristics of the 17 benchmark datasets for ordinal regression, which originate from \cite{asuncion2007uci,pascal2011}. These datasets have real ordinal targets and are standard benchmarks based on prior work \cite{gutierrez2015ordinal}.}
\label{tab:info}
\adjustbox{max width=0.99\columnwidth}{%
\begin{tabular}{lrrr}
\toprule
Dataset & Samples & Features & Classes \\
\midrule
ERA & 1000 & 4 & 9 \\
ESL & 488 & 4 & 9 \\
LEV & 1000 & 4 & 5 \\
SWD & 1000 & 10 & 4 \\
automobile & 205 & 71 & 6 \\
balance-scale & 625 & 4 & 3 \\
bondrate & 57 & 37 & 5 \\
car & 1728 & 21 & 4 \\
contact-lenses & 24 & 6 & 3 \\
eucalyptus & 736 & 91 & 5 \\
newthyroid & 215 & 5 & 3 \\
pasture & 36 & 25 & 3 \\
squash-stored & 52 & 51 & 3 \\
squash-unstored & 52 & 52 & 3 \\
tae & 151 & 54 & 3 \\
toy & 300 & 2 & 5 \\
winequality-red & 1599 & 11 & 6 \\
\bottomrule
\end{tabular}
}
\end{table}

An alternative hypothesis may be that a non-linear model would alleviate the need for coGOL, by allowing the linearly restrictive parallel hyperplanes of the standard OL approach to become more flexible with a non-linear transformation, while retaining its powerful inductive bias. We assess this hypothesis by implementing kernelized variants of all three methods: OL, GOL, and coGOL, and testing on the same datasets. 

We use the Radial Basis Function (RBF) kernel. As before, hyperparameter search was done with stratified 3-fold cross-validation and 40 iterations of Optuna. We sample $\gamma$ from a log-uniform distribution over the range $[0.01 / (2\tau_0^2), 100 / (2\tau_0^2)$] where $\tau_0$ is defined by the $1/K$ quantile of the pairwise Euclidean distances over the training set, as suggested in \cite{chapelle2005semi}.
The linear and kernel models were trained on a mix of 200 Intel Xeon Silver 4214 and Core i7-7820X CPUs.

\subsection{Deep learning experiments}

In a second set of experiments, we consider multiple datasets suitable for a deep learning approach, due to large size or non-tabular data structure. We compare deep coGOL with the standard deep OL (that is, $\beta \to \infty$), OR-CNN \cite{niu2016ordinal}, CORAL \cite{cao2019rank}, the order regularization (Reg-Ord) model from \cite{guo2021order}, as well as a cross-entropy classifier (CE). To reasonably tune the 5 hyperparameters of Reg-Ord, we try the three most common settings from their paper and choose the best performance. Where available we also include results from Liu et al. \cite{liu2018constrained}. In all experiments we tune the models using the validation set and select the best checkpoint over all epochs by validation MAE performance. We replicate each experiment 3 times and report average metrics, with the exception of the Historical Color Images Dataset (see below). Each model was trained on an NVIDIA RTX 2080 or NVIDIA RTX 6000 in under 1.5 hours. Dataset and training details follow. 

\textbf{BRFSS:}
The Behavioral Risk Factor Surveillance System (BRFSS) is an annual health survey administered by the Centers for Disease Control and Prevention. The survey collects comprehensive health and demographic information from residents across a large part of the United States. Similar to \cite{wang2019tackling}, we preprocess the 2016 BRFSS dataset to 80 features. Our prediction target is the BMI category of each respondent, which falls in 4 classes: underweight, normal weight, overweight, \& obese.

We use a 80-10-10 split for train, validation, and test sets. As the base model, we employ a 3-layer feedforward network with 100 hidden units each and ELU activation. We train the model with Adam optimizer with learning rate 1e-3 and batch size 10000 over 50 epochs. $\alpha,\beta$ were selected as 0.001, 0.01 respectively.

\textbf{AFAD:}
Following previous work~\cite{niu2016ordinal,cao2019rank}, we also consider AFAD, a large dataset that has been used for age prediction from cropped and centered facial images. We used the same subset from \cite{cao2019rank}, consisting of 165,501 face images with age range 15-40. The same test set was used, but we found that the original validation set was unrepresentative, containing a small subset of the set of all ages in the full dataset (i.e., most training set ages do not appear in validation). We thus reshuffled the training and validation sets with a new 85-15 split. We generally followed the training of their paper with modifications to improve performance. We adopted a ResNet34 as the base model, initialized using pretrained ImageNet weights. The models were trained with Adam optimizer at learning rate 5e-4 for 30 epochs and batch size 256. $\alpha,\beta$ were selected as 0.001, 0.01 respectively.

\textbf{Radio Frequency:}
The Radio Machine Learning dataset is a large collection of radio signals, both real and simulated, from 24 different channels \cite{o2018over}. Significant subsets of the channels exhibit an ordinal relationship. For example, classes QPSK, 16QAM, 32QAM, 64QAM, 128QAM, 256QAM have the same encoding approach but differ only in the modulation of amplitude to change how many bits are encoded at a time (i.e,  4, 16, 32, 64, etc. bits). In our experiment, we subset the data to the above categories to assess whether an ordinal modeling approach outperforms classification in prediction error. This subset contains 106,496 samples from each channel for a total of 638,976 samples.
Each signal is a length-1024 1-dimensional sequence with 2 channels. Adapting the method of \cite{o2018over}, our base model is a modified ResNet with 6 stacks, followed by 3 FC layers (details in Appendix). We train the model with Adam optimizer with learning rate 5e-4, batch size 2048, trained for 50 epochs. We use an 80-20 train-test split with further 10\% of training used for the validation set. $\alpha,\beta$ were selected as 0.01, 0.05 respectively.

\textbf{Historical Color Images:}
The Historical Color Image Dataset (HCID) is a collection of 1,325 historical color photographs from the 1930s to the 1970s \cite{palermo2012dating}. Because of the changing technologies behind photography, the target of this task is predicting the decade in which each photograph was taken. The dataset spans 5 decades with 275 images each. Following recent work in \cite{liu2018constrained}, we adopt a 210-5-50 stratified split. We include their results as a benchmark. Because the original splits were not available and due to the small dataset size, we replicate our experiment over 5 random train-valid-test splits for fair comparison.
We kept our training method similar to \cite{liu2018constrained} but with modifications due to different model characteristics. The base model was a VGG13 with batch normalization, pretrained on ImageNet. We trained with Adam optimizer with learning rate 5e-5, batch size 128, for 20 epochs. $\alpha,\beta$ were selected as 0.01, 0.05 respectively.

\section{Results} \label{sec:results}

\begin{table*}[t]
\caption{Model results on 17 ordinal regression benchmark datasets using Linear and RBF kernel models. Best linear results shown in \textbf{bold}. Datasets that are {\ul underlined} indicate that our Linear coGOL performs better than any kernelized model in MAE, showing that kernelization is not sufficient to tackle the learning constraints posed by OL. Our linear coGOL model shows a statistically significant improvement over the ordinal logistic (OL) baseline. The five datasets where RBF kernel performs better than linear in MAE are denoted with ``*'' and are the only cases where we highlight \textbf{best} and \textit{second best} RBF results. Our RBF kernel coGOL is best or second best in these cases.}
\label{tab:benchmarks}
\centering
\adjustbox{max width=0.99\textwidth}{%
\begin{tabular}{@{}llcccccclcccccc@{}}
\toprule
                                      &                           & \multicolumn{3}{c}{Linear}                       & \multicolumn{3}{c}{RBF}                          &                                        & \multicolumn{3}{c}{Linear}                       & \multicolumn{3}{c}{RBF}                          \\  \cmidrule(lr){3-5}\cmidrule(lr){6-8} \cmidrule(l){10-12} \cmidrule(l){13-15} 
\multicolumn{1}{c}{Dataset}           & \multicolumn{1}{c}{Model} & MAE            & MSE            & Acc            & MAE            & MSE            & Acc            & \multicolumn{1}{c}{Dataset}            & MAE            & MSE            & Acc            & MAE            & MSE            & Acc            \\ \midrule
\multirow{3}{*}{{\ul ERA}}            & OL                        & 1.210          & 2.533          & 0.256          & 1.212          & 2.512          & 0.252          & \multirow{3}{*}{{\ul eucalyptus}}      & 0.396          & 0.457          & 0.634          & 0.516          & 0.712          & 0.560          \\
                                      & GOL                       & \textbf{1.192} & \textbf{2.509} & \textbf{0.264} & 1.224          & 2.788          & 0.240          &                                        & 0.404          & 0.502          & 0.636          & 0.500          & 0.723          & 0.582          \\
                                      & coGOL                     & \textbf{1.192} & 2.514 & \textbf{0.264}          & 1.216          & 2.584          & 0.264          &                                        & \textbf{0.386} & \textbf{0.456} & \textbf{0.645} & 0.462          & 0.603          & 0.598          \\ \midrule
\multirow{3}{*}{\ul ESL}                 & OL                     & 0.310          & 0.345          & 0.705          & 0.303          & 0.328          & 0.713          & \multirow{3}{*}{{\ul newthyroid}}         & \textbf{0.030} & \textbf{0.030} & \textbf{0.970} & 0.037          & 0.037          & 0.963          \\
                                      & GOL                       & 0.309          & 0.340          & 0.706          & 0.320          & 0.344          & 0.689          &                                        & 0.033          & 0.033          & 0.967          & 0.037          & 0.037          & 0.963          \\
                                      & coGOL                     & \textbf{0.293} & \textbf{0.324} & \textbf{0.722} & 0.311          & 0.336          & 0.697          &                                        & \textbf{0.030} & \textbf{0.030} & \textbf{0.970} & 0.037          & 0.037          & 0.963          \\ \midrule
\multirow{3}{*}{LEV*}                 & OL                        & \textbf{0.416} & \textbf{0.490} & \textbf{0.618} & 0.414          & 0.496          & 0.620          & \multirow{3}{*}{{\ul pasture}}         & 0.333          & 0.333          & 0.667          & 0.444          & 0.444          & 0.556          \\
                                      & GOL                       & 0.434          & 0.509          & 0.601          & \textit{0.410} & \textit{0.492} & \textbf{0.630} &                                        & \textbf{0.283} & \textbf{0.283} & \textbf{0.717} & 0.444          & 0.444          & 0.556          \\
                                      & coGOL                     & 0.421          & 0.494          &     0.613      & \textbf{0.406} & \textbf{0.482} & \textit{0.628} &                                        & 0.328          & 0.328          & 0.672          & 0.444          & 0.444          & 0.556          \\ \midrule
\multirow{3}{*}{{\ul SWD}}            & OL                        & 0.448          & 0.484          & 0.570          & 0.436          & 0.474          & 0.582          & \multirow{3}{*}{{\ul squash-stored}}   & 0.416          & 0.416          & 0.437          & 0.462          & 0.462          & 0.538          \\
                                      & GOL                       & 0.435          & \textbf{0.464} & 0.579          & 0.438          & 0.472          & 0.580          &                                        & 0.381          & 0.430          & \textbf{0.643} & 0.462          & 0.538          & 0.615          \\
                                      & coGOL                     & \textbf{0.433} & \textbf{0.464} & \textbf{0.582} & 0.438          & 0.470          & 0.580          &                                        & \textbf{0.378} & \textbf{0.413} & 0.640          & 0.385          & 0.423          & 0.615          \\ \midrule
\multirow{3}{*}{{\ul automobile}}     & OL                        & 0.454          & 0.601          & 0.607          & 0.596          & 0.808          & 0.500          & \multirow{3}{*}{{\ul squash-unstored}} & 0.289          & \textbf{0.289} & 0.711          & 0.308          & 0.308          & 0.692          \\
                                      & GOL                       & \textbf{0.369} & \textbf{0.513} & 0.696          & 0.404          & 0.558          & 0.673          &                                        & 0.282          & 0.297          & 0.725          & 0.308          & 0.308          & 0.692          \\
                                      & coGOL                     & 0.379          & 0.590          & \textbf{0.714} & 0.404          & 0.615          & 0.673          &                                        & \textbf{0.275} & \textbf{0.289} & \textbf{0.733}          & 0.308          & 0.308          & 0.692          \\ \midrule
\multirow{3}{*}{balance-scale*}       & OL                        & \textbf{0.107} & \textbf{0.130} & \textbf{0.905} & \textbf{0.013} & \textbf{0.013} & \textbf{0.987} & \multirow{3}{*}{{tae*}}                & 0.614          & 0.752          & 0.455          & \textbf{0.553}          & \textbf{0.724}          & 0.500          \\
                                      & GOL                       & \textbf{0.107} & 0.131          & \textbf{0.905} & 0.019          & 0.019          & 0.981          &                                        & \textbf{0.575} & 0.761          & \textbf{0.518} & 0.579          & \textit{0.776 }         & \textit{0.513}          \\
                                      & coGOL                     & \textbf{0.107} & 0.131          & \textbf{0.905} & \textit{0.016} & \textit{0.016} & \textit{0.984} &                                        & 0.580          & \textbf{0.741} & 0.500          & \textit{0.566}          & \textit{0.776}          & \textbf{0.526}          \\ \midrule
\multirow{3}{*}{{\ul bondrate}}       & OL                        & 0.549          & 0.767          & 0.556          & 0.600          & 1.133          & 0.600          & \multirow{3}{*}{{toy*}}                & 0.953          & 1.464          & 0.287          & \textbf{0.020} & \textbf{0.020} & \textbf{0.980} \\
                                      & GOL                       & 0.564          & 0.818          & 0.556          & 0.667          & 1.200          & 0.533          &                                        & 0.899          & 1.325          & 0.294          & 0.053          & 0.053          & 0.947          \\
                                      & coGOL                     & \textbf{0.536} & \textbf{0.744} & \textbf{0.562} & 0.633          & 1.133          & 0.567          &                                        & \textbf{0.898} & \textbf{1.324} & \textbf{0.295} & \textit{0.040}          & \textit{0.040}          & \textit{0.960}          \\ \midrule
\multirow{3}{*}{{\ul car}}            & OL                        & 0.082          & 0.083          & 0.919          & 0.412          & 0.587          & 0.676          & \multirow{3}{*}{winequality-red*}      & 0.443          & 0.515          & 0.592          & \textit{0.435}          & \textit{0.540}          & \textbf{0.611}          \\
                                      & GOL                       & \textbf{0.073} & \textbf{0.078} & \textbf{0.930} & 0.339          & 0.426          & 0.701          &                                        & \textbf{0.436} & \textbf{0.511} & \textbf{0.600} & 0.446          & 0.556          & 0.605 \\
                                      & coGOL                     & \textbf{0.073} & \textbf{0.078} & \textbf{0.930} & 0.325          & 0.411          & 0.712          &                                        & 0.438          & 0.516          & 0.598          & \textbf{0.433} & \textbf{0.524} & \textit{0.608}          \\ \midrule
\multirow{3}{*}{{\ul contact-lenses}} & OL                        & 0.428          & 0.601          & 0.659          & 0.667          & 0.917          & 0.500          & \multirow{3}{*}{}                      &                &                &                &                &                &                \\
                                      & GOL                       & \textbf{0.384} & \textbf{0.572} & \textbf{0.710} & 0.583          & 0.833          & 0.500          &                                        &                &                &                &                &                &                \\
                                      & coGOL                     & 0.406          & 0.623          & 0.703          & 0.667          & 0.917          & 0.500          &                                        &                &                &                &                &                &                \\ \bottomrule
\end{tabular}
}
\end{table*}


\subsection{Linear models}
Recalling that coGOL enables a continuous interpolation between the all-thresholds ordinal logit (OL) and the generalized ordinal logit (GOL) models, we compare coGOL to these two models. The OL model is a standard approach as described in \cite{rennie2005loss,pedregosa2017consistency} so we view it as the baseline. Because we introduced generalized coefficients to the all-thresholds OL model, the GOL results serve as an intermediate result. We calculated the mean absolute error (MAE), mean squared error (MSE), and zero-one error (accuracy) over 30 test set splits.

Evaluation metrics are shown in \autoref{tab:benchmarks}. coGOL is competitive with or better than both comparison models, achieving lowest MAE in 11 out of 17 datasets, lowest MSE in 10, and highest accuracy in 11. In contrast, OL has the lowest MAE in 3, lowest MSE in 4, and highest accuracy in 3. Using the Wilcoxon signed rank test to compare coGOL with OL, we find that the difference in MAE is highly significant, with $p$=0.0004. The difference in accuracy is also highly significant, with $p$=0.0001, while MSE is significant at the $\alpha=0.1$ level at $p$=0.066.

Consider a data generation process that obeys the parallel threshold assumption. Here, OL is sample-efficient and learns the optimal model. While GOL can learn the same model, it may overfit if data variance is high, giving worse generalization. The regularization parameter on coGOL can help to control the model flexibility in this case by restricting the deviation between the different weights. On the other hand, when the data does not follow the parallel assumption, the OL model will be too restrictive. While GOL may be the best choice in low-noise data distributions, coGOL with weak regularization may still be better in the presence of sufficient data noise. Experimentally, we observe that in 5 out of the 6 largest datasets, coGOL outperforms OL. coGOL also outperforms both GOL in \textit{bondrate}, \textit{squash-stored}, and \textit{squash-unstored}, 3 small datasets with high feature-to-sample ratio. On some datasets such as \textit{balance-scale} all models perform the same, indicating that the OL assumptions are satisfactory for the data distribution.

\subsection{Kernel models}

One may hypothesize that the non-linearity introduced by a kernelization would allow the OL model to attain the expressiveness of coGOL. However, a kernel model introduces new hyper-parameters and variance increases due to the additional degrees of freedom in the model. Because ordinal models are most valuable in low-sample size problems where the ordinal hypothesis serves as a strong inductive bias, this increases the difficulty of the task in a non-trivial manner. Indeed, \autoref{tab:benchmarks} shows that the linear coGOL model performs better than any kernelized variant in 12 datasets. Of the remaining 5, only two (balance-scale and toy) show significant decreased error, indicating that the classes were not linearly separable. The last three datasets only show modest improvement over the linear models. This suggests that kernelization is not sufficient to overcome the constraints of the OL model when factoring in the increased model complexity and variance in results. This highlights that the significant performance advantage that coGOL provides in the linear case is meaningful.


\subsection{Deep models}
Following the experimental procedure in \autoref{sec:experiments}, we evaluated the MAE and root mean square error (RMSE) of coGOL compared with multiple benchmarks. Results are shown in \autoref{tab:deep_results}. Across all datasets, we find that generalized coefficient models (coGOL, OR-CNN, Reg-ORD) uniformly improve performance compared to the threshold approach (OL, CORAL). coGOL consistently has lower MAE and either lowest or second-lowest RMSE. In particular there is a dramatic improvement of MAE and RMSE on HCID, indicating the standout effect of neighboring-class regularization in smaller deep learning datasets. coGOL also uniformly outperforms naive classification (CE), further highlighting the benefit of using specialized ordinal regression approaches. 

\begin{figure}[!ht]
    \centering
    \adjustbox{max width=0.9\columnwidth}{%
\begin{tikzpicture}

\definecolor{color0}{rgb}{0.12156862745098,0.466666666666667,0.705882352941177}
\definecolor{color1}{rgb}{1,0.498039215686275,0.0549019607843137}
\definecolor{color2}{rgb}{0.172549019607843,0.627450980392157,0.172549019607843}
\definecolor{color3}{rgb}{0.83921568627451,0.152941176470588,0.156862745098039}
\definecolor{color4}{rgb}{0.580392156862745,0.403921568627451,0.741176470588235}

\begin{axis}[
legend cell align={left},
legend style={fill opacity=0.8, draw opacity=1, text opacity=1, draw=none},
tick align=outside,
tick pos=left,
x grid style={white!69.0196078431373!black},
xlabel={Epoch},
xmin=-1.45, xmax=30.45,
xtick style={color=black},
y grid style={white!69.0196078431373!black},
ylabel={Validation MAE},
ymin=2.94885, ymax=7.40015,
ytick style={color=black},
height=0.99\columnwidth,
width=0.99\columnwidth,
]
\addplot [semithick, color0]
table {%
0 3.378
1 3.725
2 3.252
3 3.189
4 3.375
5 3.175
6 3.351
7 3.213
8 3.215
9 3.269
10 3.207
11 3.197
12 3.204
13 3.234
14 3.263
15 3.332
16 3.315
17 3.275
18 3.456
19 3.392
20 3.279
21 3.318
22 3.349
23 3.246
24 3.416
25 3.32
26 3.312
27 3.302
28 3.305
29 3.351
};
\addlegendentry{coGOL}
\addplot [semithick, color1]
table {%
0 3.593
1 3.367
2 3.263
3 3.482
4 3.252
5 3.204
6 3.251
7 3.153
8 3.262
9 3.231
10 3.228
11 3.178
12 3.229
13 3.303
14 3.486
15 3.312
16 3.469
17 3.219
18 3.401
19 3.317
20 3.273
21 3.227
22 3.251
23 3.352
24 3.316
25 3.536
26 3.272
27 3.409
28 3.307
29 3.379
};
\addlegendentry{OR-CNN}
\addplot [semithick, color2]
table {%
0 7.236
1 6.201
2 5.626
3 4.771
4 4.326
5 3.896
6 3.985
7 3.882
8 3.856
9 3.661
10 3.658
11 3.622
12 3.524
13 3.407
14 3.442
15 3.416
16 3.425
17 3.34
18 3.317
19 3.266
20 3.467
21 3.302
22 3.367
23 3.341
24 3.391
25 3.426
26 3.324
27 3.33
28 3.299
29 3.26
};
\addlegendentry{OL}
\addplot [semithick, color3]
table {%
0 7.208
1 6.101
2 5.397
3 4.857
4 4.29
5 4.096
6 4.001
7 4.679
8 3.641
9 3.671
10 3.586
11 3.513
12 3.489
13 3.493
14 3.445
15 3.394
16 3.364
17 3.409
18 3.357
19 3.375
20 3.41
21 3.308
22 3.385
23 3.288
24 3.325
25 3.324
26 3.305
27 3.312
28 3.243
29 3.265
};
\addlegendentry{CORAL}
\addplot [semithick, color4]
table {%
0 3.781
1 3.678
2 3.56
3 3.571
4 3.444
5 3.416
6 3.456
7 3.578
8 3.488
9 3.495
10 3.619
11 3.618
12 3.41
13 3.57
14 3.519
15 3.555
16 3.648
17 3.658
18 3.743
19 3.704
20 3.635
21 3.693
22 3.567
23 3.971
24 3.604
25 3.644
26 3.764
27 3.885
28 3.722
29 3.707
};
\addlegendentry{CE}
\end{axis}

\end{tikzpicture}
    }
    \caption{MAE by number of epochs for each model on the AFAD dataset.
    }
    \label{fig:valid}
\end{figure}
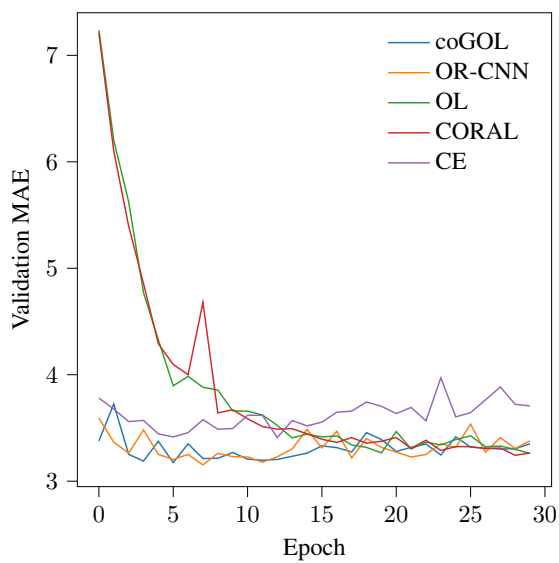

Small values for $\beta$ often work best in the deep learning setting. For this reason, other generalized models (OR-CNN, Reg-Ord) can be competitive with coGOL. Compared to the linear benchmarks, the challenge of the deep learning datasets involve learning meaningful representations rather than dealing with data noise. This tends to favor flexible models. Furthermore, with increasing dataset complexity, it becomes less likely that the relation between features extracted by the base neural network and the ordinal class satisfies the parallel assumption. For example, we would expect for HCID that changes in camera technology were complex and nonlinear over time, so the same features that separate historical images from one decade may not useful in the next. coGOL's flexibility lends it a better inductive prior to learn these complex decision boundaries, while enforcing neighbor-class smoothness encourages the embedding space to remain organized. In contrast, the more rigid OL models would likely need to learn a further transformation to align the embeddings linearly. This is further supported by training analysis in \autoref{fig:valid} showing validation error as a function of epoch number. We observe that the generalized models reach optimal validation loss in under 10 epochs, while the OL-based models improve much more slowly.


\begin{table}[t]
    \caption{The performance of our model coGOL, compared to benchmarks on four deep learning datasets. The results highlight that OR-CNN and CORAL are special cases of our approach. Best results in \textbf{bold}, second best in \textit{italics}. }
    \label{tab:deep_results}
    \centering
    \adjustbox{max width=\columnwidth}{%
\begin{tabular}{@{}ccccccccc@{}}
\toprule
           & \multicolumn{2}{c}{BRFSS} & \multicolumn{2}{c}{AFAD} & \multicolumn{2}{c}{RF} & \multicolumn{2}{c}{HCID} \\ 
             \cmidrule(lr){2-3}          \cmidrule(lr){4-5}         \cmidrule(lr){6-7}       \cmidrule(lr){8-9}
           & MAE         & RMSE        & MAE         & RMSE       & MAE        & RMSE      & MAE         & RMSE       \\ \midrule
coGOL      & \textbf{0.580} & \textbf{0.825} & \textbf{3.209} & \textit{4.454} & \textbf{0.615} & \textit{1.338} & \textbf{0.674} & \textbf{1.033} \\
OL         & 0.592          &  0.839          & 3.310         & 4.560 & 0.953          &     1.384 & 1.215          & 1.690          \\
OR-CNN & \textit{0.581} & \textbf{0.825 }            &    3.229            & 4.483             & \textit{0.616}             & 1.367            & 0.707           & 1.087             \\
CORAL      & 0.591          & 0.836     &     3.300          &   4.519          & 0.945          & 1.400          & 1.195          & 1.667          \\
Reg-Ord & \textit{0.581} & \textit{0.828} & \textit{3.217} & \textbf{4.440} & 0.619 & \textbf{1.247} & \textit{0.689} & \textit{1.066} \\
CE         & 0.618          & 0.918          & 3.439 & 4.871          & 0.654 & 1.500          & 0.750 &  1.193 \\
Liu et al. & --             & --             & --             & --             & --             & --             & 0.82           & --             \\ \bottomrule
\end{tabular}
    }

\end{table}

We also note that OR-CNN and Reg-Ord were only formulated for deep learning. In particular the softmax loss function formulation of OR-CNN and the large hyperparameter space of Reg-Ord cause significant training difficulty for the linear model benchmark. Therefore our coGOL model improves on other GOL-based models in deep learning and is novel to the linear case.

Finally, the kernel and deep results indicate that a non-linear mapping from the original feature space is not sufficient to overcome the restrictions of the OL threshold model. In other words, it does not appear efficient to learn a mapping of the data points to a space that can be separated by parallel hyperplanes.

\section{Conclusion} \label{sec:conclusion}

We have developed the continuously generalized ordinal logistic model coGOL. The most common linear and deep ordinal models can be derived as special cases of our approach, which introduces a single hyper parameter $\beta$ to balance model flexibility vs. helpful inductive bias. Our approach unifies the most common linear and deep models into a single framework, while also introducing a spectrum of ordinal models not previously developed. Our coGOL approach obtains comparable or better results across linear and deep learning based models, showing a statistically significant improvement.






\bibliographystyle{siamplain}
\bibliography{ijcai21}

\begin{thebibliography}{10}

\bibitem{agrawal2018rewriting}
{\sc A.~Agrawal, R.~Verschueren, S.~Diamond, and S.~Boyd}, {\em A rewriting
  system for convex optimization problems}, Journal of Control and Decision, 5
  (2018), pp.~42--60.

\bibitem{agresti2013modeling}
{\sc A.~Agresti}, {\em Modeling ordinal categorical data}, University of
  Florida, Department of Statistics,  (2013).

\bibitem{akiba2019optuna}
{\sc T.~Akiba, S.~Sano, T.~Yanase, T.~Ohta, and M.~Koyama}, {\em Optuna: A
  next-generation hyperparameter optimization framework}, in Proceedings of the
  25th ACM SIGKDD international conference on knowledge discovery \& data
  mining, 2019, pp.~2623--2631.

\bibitem{Alpaydin:1999:CTC:339993.339999}
{\sc E.~Alpaydin}, {\em {Combined 5 × 2 cv F Test for Comparing Supervised
  Classification Learning Algorithms}}, Neural Comput., 11 (1999),
  pp.~1885--1892, \url{https://doi.org/10.1162/089976699300016007},
  \url{http://dx.doi.org/10.1162/089976699300016007}.

\bibitem{ananth1997regression}
{\sc C.~V. Ananth and D.~G. Kleinbaum}, {\em Regression models for ordinal
  responses: a review of methods and applications.}, International journal of
  epidemiology, 26 (1997), pp.~1323--1333.

\bibitem{asuncion2007uci}
{\sc A.~Asuncion and D.~Newman}, {\em Uci machine learning repository}, 2007.

\bibitem{JMLR:v17:benavoli16a}
{\sc A.~Benavoli, G.~Corani, and F.~Mangili}, {\em {Should We Really Use
  Post-Hoc Tests Based on Mean-Ranks?}}, Journal of Machine Learning Research,
  17 (2016), pp.~1--10, \url{http://jmlr.org/papers/v17/benavoli16a.html}.

\bibitem{Bengio:2004:NUE:1005332.1044695}
{\sc Y.~Bengio and Y.~Grandvalet}, {\em {No Unbiased Estimator of the Variance
  of K-Fold Cross-Validation}}, Journal of Machine Learning Research, 5 (2004),
  pp.~1089--1105, \url{http://dl.acm.org/citation.cfm?id=1005332.1044695}.

\bibitem{Blum:1999:BHB:307400.307439}
{\sc A.~Blum, A.~Kalai, and J.~Langford}, {\em {Beating the Hold-out: Bounds
  for K-fold and Progressive Cross-validation}}, in Proceedings of the Twelfth
  Annual Conference on Computational Learning Theory, COLT '99, New York, NY,
  USA, 1999, ACM, pp.~203--208, \url{https://doi.org/10.1145/307400.307439},
  \url{http://doi.acm.org/10.1145/307400.307439}.

\bibitem{cao2019rank}
{\sc W.~Cao, V.~Mirjalili, and S.~Raschka}, {\em Rank-consistent ordinal
  regression for neural networks}, arXiv preprint arXiv:1901.07884,  (2019).

\bibitem{chapelle2005semi}
{\sc O.~Chapelle and A.~Zien}, {\em Semi-supervised classification by low
  density separation.}, in AISTATS, vol.~2005, Citeseer, 2005, pp.~57--64.

\bibitem{cheng2008neural}
{\sc J.~Cheng, Z.~Wang, and G.~Pollastri}, {\em A neural network approach to
  ordinal regression}, in 2008 IEEE International Joint Conference on Neural
  Networks (IEEE World Congress on Computational Intelligence), IEEE, 2008,
  pp.~1279--1284.

\bibitem{chu2005gaussian}
{\sc W.~Chu, Z.~Ghahramani, and C.~K. Williams}, {\em Gaussian processes for
  ordinal regression.}, Journal of machine learning research, 6 (2005).

\bibitem{chu2007support}
{\sc W.~Chu and S.~S. Keerthi}, {\em Support vector ordinal regression}, Neural
  computation, 19 (2007), pp.~792--815.

\bibitem{Demsar:2006:SCC:1248547.1248548}
{\sc J.~Dem{\v{s}}ar}, {\em {Statistical Comparisons of Classifiers over
  Multiple Data Sets}}, Journal of Machine Learning Research, 7 (2006),
  pp.~1--30, \url{http://dl.acm.org/citation.cfm?id=1248547.1248548}.

\bibitem{cvxpy}
{\sc S.~Diamond and S.~Boyd}, {\em {CVXPY}: A {P}ython-embedded modeling
  language for convex optimization}, Journal of Machine Learning Research,
  (2016), \url{http://stanford.edu/~boyd/papers/pdf/cvxpy_paper.pdf}.
\newblock To appear.

\bibitem{diaz2019soft}
{\sc R.~Diaz and A.~Marathe}, {\em Soft labels for ordinal regression}, in
  Proceedings of the IEEE/CVF Conference on Computer Vision and Pattern
  Recognition, 2019, pp.~4738--4747.

\bibitem{Dietterich:1998:AST:303222.303237}
{\sc T.~G. Dietterich}, {\em {Approximate Statistical Tests for Comparing
  Supervised Classification Learning Algorithms}}, Neural Comput., 10 (1998),
  pp.~1895--1923, \url{https://doi.org/10.1162/089976698300017197},
  \url{http://dx.doi.org/10.1162/089976698300017197}.

\bibitem{frank2001simple}
{\sc E.~Frank and M.~Hall}, {\em A simple approach to ordinal classification},
  in European conference on machine learning, Springer, 2001, pp.~145--156.

\bibitem{guo2021order}
{\sc T.~Guo, H.~Zhang, B.~Yoo, Y.~Liu, Y.~Kwak, and J.-J. Han}, {\em Order
  regularization on ordinal loss for head pose, age and gaze estimation}, in
  Proceedings of the AAAI Conference on Artificial Intelligence, vol.~35, 2021,
  pp.~1496--1504.

\bibitem{gutierrez2015ordinal}
{\sc P.~A. Guti{\'e}rrez, M.~Perez-Ortiz, J.~Sanchez-Monedero,
  F.~Fernandez-Navarro, and C.~Hervas-Martinez}, {\em Ordinal regression
  methods: survey and experimental study}, IEEE Transactions on Knowledge and
  Data Engineering, 28 (2015), pp.~127--146.

\bibitem{levi2015age}
{\sc G.~Levi and T.~Hassner}, {\em Age and gender classification using
  convolutional neural networks}, in Proceedings of the IEEE conference on
  computer vision and pattern recognition workshops, 2015, pp.~34--42.

\bibitem{li2006ordinal}
{\sc L.~Li and H.-T. Lin}, {\em Ordinal regression by extended binary
  classification}, Advances in neural information processing systems, 19
  (2006), pp.~865--872.

\bibitem{lin2008ordinal}
{\sc H.-T. Lin}, {\em From ordinal ranking to binary classification},
  California Institute of Technology, 2008.

\bibitem{lin2012reduction}
{\sc H.-T. Lin and L.~Li}, {\em Reduction from cost-sensitive ordinal ranking
  to weighted binary classification}, Neural Computation, 24 (2012),
  pp.~1329--1367.

\bibitem{liu2018constrained}
{\sc Y.~Liu, A.~Wai Kin~Kong, and C.~Keong~Goh}, {\em A constrained deep neural
  network for ordinal regression}, in Proceedings of the IEEE Conference on
  Computer Vision and Pattern Recognition, 2018, pp.~831--839.

\bibitem{mathieson1996ordinal}
{\sc M.~J. Mathieson}, {\em Ordinal models for neural networks}, in Proc. 3rd
  Int. Conf. Neural Netw. Capital Markets, 1996, pp.~523--536.

\bibitem{mccullagh1980regression}
{\sc P.~McCullagh}, {\em Regression models for ordinal data}, Journal of the
  Royal Statistical Society: Series B (Methodological), 42 (1980),
  pp.~109--127.

\bibitem{niu2016ordinal}
{\sc Z.~Niu, M.~Zhou, L.~Wang, X.~Gao, and G.~Hua}, {\em Ordinal regression
  with multiple output cnn for age estimation}, in Proceedings of the IEEE
  conference on computer vision and pattern recognition, 2016, pp.~4920--4928.

\bibitem{o2018over}
{\sc T.~J. O’Shea, T.~Roy, and T.~C. Clancy}, {\em Over-the-air deep learning
  based radio signal classification}, IEEE Journal of Selected Topics in Signal
  Processing, 12 (2018), pp.~168--179.

\bibitem{palermo2012dating}
{\sc F.~Palermo, J.~Hays, and A.~A. Efros}, {\em Dating historical color
  images}, in European Conference on Computer Vision, Springer, 2012,
  pp.~499--512.

\bibitem{pascal2011}
{\sc PASCAL}, {\em Pascal (pattern analysis, statistical modelling and
  computational learning) machine learning benchmarks repository}.
\newblock \url{http://mldata.org}, 2011.

\bibitem{pedregosa2017consistency}
{\sc F.~Pedregosa, F.~Bach, and A.~Gramfort}, {\em On the consistency of
  ordinal regression methods}, The Journal of Machine Learning Research, 18
  (2017), pp.~1769--1803.

\bibitem{rennie2005loss}
{\sc J.~D. Rennie and N.~Srebro}, {\em Loss functions for preference levels:
  Regression with discrete ordered labels}, in Proceedings of the IJCAI
  multidisciplinary workshop on advances in preference handling, vol.~1, Kluwer
  Norwell, MA, 2005.

\bibitem{rothe2015dex}
{\sc R.~Rothe, R.~Timofte, and L.~Van~Gool}, {\em Dex: Deep expectation of
  apparent age from a single image}, in Proceedings of the IEEE international
  conference on computer vision workshops, 2015, pp.~10--15.

\bibitem{Varma2006}
{\sc S.~Varma and R.~Simon}, {\em {Bias in error estimation when using
  cross-validation for model selection}}, 2006,
  \url{https://doi.org/10.1186/1471-2105-7-91}.

\bibitem{waegeman2009ensemble}
{\sc W.~Waegeman, L.~Boullart, et~al.}, {\em An ensemble of weighted support
  vector machines for ordinal regression}, International Journal of Computer
  Systems Science and Engineering, 3 (2009), pp.~47--51.

\bibitem{wang2017nonparallel}
{\sc H.~Wang, Y.~Shi, L.~Niu, and Y.~Tian}, {\em Nonparallel support vector
  ordinal regression}, IEEE transactions on cybernetics, 47 (2017),
  pp.~3306--3317.

\bibitem{wang2019tackling}
{\sc L.~Wang and D.~Zhu}, {\em Tackling multiple ordinal regression problems:
  Sparse and deep multi-task learning approaches}, arXiv preprint
  arXiv:1907.12508,  (2019).

\bibitem{Wilcoxon1945}
{\sc F.~Wilcoxon}, {\em {Individual Comparisons by Ranking Methods}},
  Biometrics Bulletin, 1 (1945), p.~80, \url{https://doi.org/10.2307/3001968},
  \url{http://www.jstor.org/stable/10.2307/3001968?origin=crossref}.

\bibitem{williams2016understanding}
{\sc R.~Williams}, {\em Understanding and interpreting generalized ordered
  logit models}, The Journal of Mathematical Sociology, 40 (2016), pp.~7--20.

\end{thebibliography}

\newpage
\appendix
\onecolumn

\section{Toy Example} \label{sec:toy}

\begin{figure}[!h]
    \centering
    \adjustbox{width=\columnwidth}{%
    \input{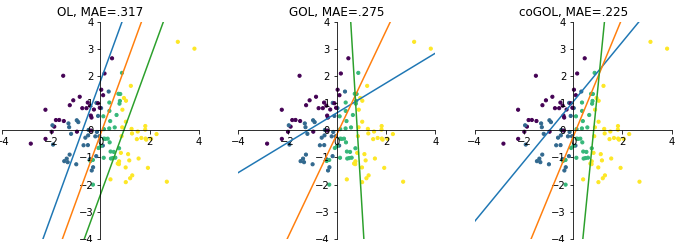}
    }
    \adjustbox{width=\columnwidth}{%
    \input{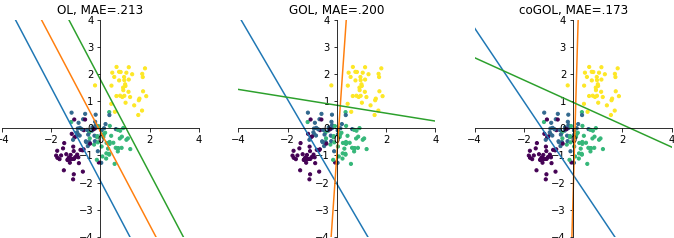}
    }
    \caption{Two toy datasets demonstrating how dramatically the hyperplanes learned can change between OL, GOL, and our coGOL. Each row shows a different dataset, and each column a different loss approach. 
    }
    \label{fig:synth1}
\end{figure}

As a small visual demonstration see \autoref{fig:synth1} for two examples on how the hyperplanes learned can change significantly by changing the ordinal loss used. In all cases the ordinal assumption is true, but the limited data available makes it challenging to learn good solutions. coGOL allows us to obtain a balance between the data efficency of OL and the expressiveness of GOL. 

\section{Comparison of threshold losses}
The following derivations are standard and can be found in their respective citations among other work. We repeat them here as references to the reader and to facilitate comparison.

\subsection{Cumulative logit loss}
From \cite{ananth1997regression,agresti2013modeling}, the cumulative logit model has $$\mathbb{P}(y \leq j | x) = \frac{\exp(\theta_j - w^\top x)}{1 + \exp(\theta_j - w^\top x)} = \sigma(\theta_j - w^\top x) $$ using $\sigma$ to represent the sigmoid function for simplicity. Then the probability of being in class $j$ is $$ P(y = j | x) = P(y \leq j | x ) - P(y \leq j - 1 | x) = \sigma(\theta_j - w^\top x) - \sigma(\theta_{j-1} - w^\top x)$$
Then the negative log-likelihood for a given sample $x$ with ordinal label $y$ among ordinal classes $1 \prec 2 \prec \ldots \prec k$ is
$$ \ell(y,f(x)) = - \log (\sigma(\theta_y - w^\top x) - \sigma(\theta_{y-1} - w^\top x))$$
with the edge cases $\theta_0=-\infty$ and $\theta_k=\infty$.

\subsection{Immediate-threshold loss}

Following the derivations of \cite{pedregosa2017consistency} and our Section 3, suppose we use multiclass accuracy as the scoring rule for our ordinal model: $\ell(y, x) = \llbracket y = x \rrbracket$. Then we can rewrite as
$$ \ell(y, f(x)) = \llbracket g_{y-1}(x) \geq 0 \rrbracket + \llbracket g_{y}(x) < 0 \rrbracket $$
for which we substitute our surrogate loss $\varphi(x) = \log(1 + e^{-x})$:
$$ \ell(y, f(x)) = \varphi(-g_{y-1}(x)) + \varphi(g_{y}(x)) $$
Noting that $\varphi(x) = -\log \sigma(x)$ and $\log\sigma(-x) = \log(1 - \sigma(x))$, we get
\begin{align*}
        \ell(y, f(x)) &= -\Big[ \log\sigma( \theta_y - w^\top x ) + \log\sigma(-(\theta_{y-1} - w^\top x))  \Big]\\
        &= -\Big[ \log\sigma( \theta_y - w^\top x ) + \log(1 - \sigma(\theta_{y-1} - w^\top x))  \Big]
\end{align*}
which highlights the similarity with the cumulative logit loss.

\subsection{All-thresholds loss}

As presented in Section 3,
\begin{align*}
    \ell(y, f(\boldsymbol{x})) &= \sum_{j=1}^{y-1} \varphi(-g_j(x)) + \sum_{j=y}^{k-1} \varphi(g_j(x))\\
    &= \sum_{j=1}^{y-1} \varphi(-(\theta_{j} - w^\top x)) + \sum_{j=y}^{k-1} \varphi(\theta_{j} - w^\top x)\\
    &= -\Big[  \sum_{j=1}^{y-1} \log(1 - \sigma(\theta_j - w^\top x)) + \sum_{j=y}^{k-1} \log\sigma(\theta_j - w^\top x)      \Big]
\end{align*}

\newpage
\section{Empirical comparison}
The 0-1 loss formula $$ \ell(y, f(x)) = \llbracket g_{y-1}(x) \geq 0 \rrbracket + \llbracket g_{y}(x) < 0 \rrbracket $$
indicates that in the absence of a standard $w^\top x$ that is partitioned using $k-1$ monotonic thresholds $\theta_j$ (OL), the 0-1 loss does not constrain the behavior of the classes outside the immediate neighbors of a sample. This alone is suggestive that generalized coefficients require the all-thresholds loss. This is supported by the following experimental results comparing all-thresholds and immediate-threshold over the 17 benchmark datasets.

\begin{figure}[!h]
    \centering
    \includegraphics[width=0.9\textwidth]{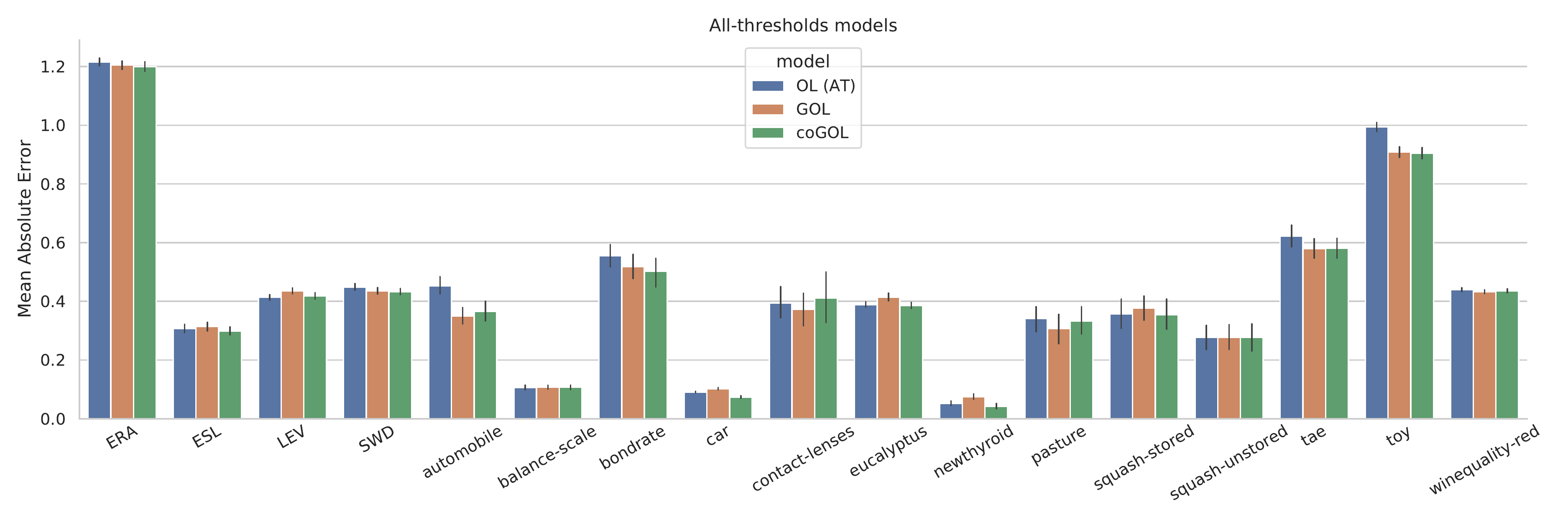}
    \caption{Comparison of coGOL, GOL, and OL for the all-thresholds loss. This is the result presented in table form in the main text.}
    \label{fig:my_label}
\end{figure}

\begin{figure}[!h]
    \centering
    \includegraphics[width=0.9\textwidth]{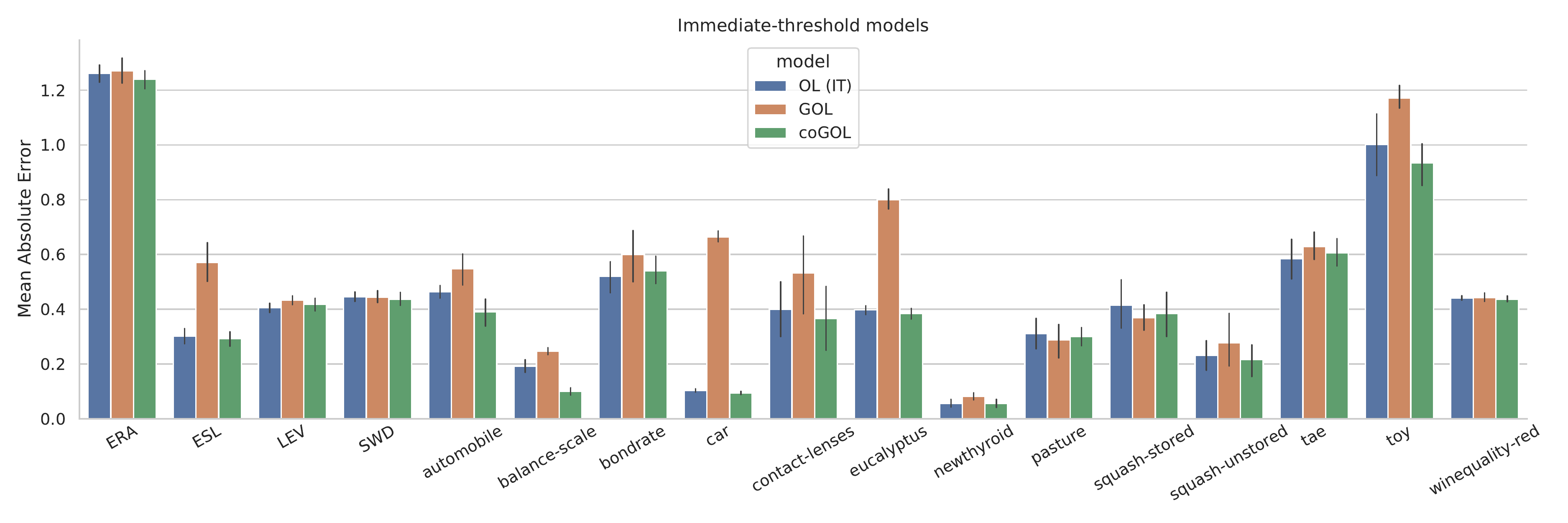}
    \caption{Comparison of coGOL, GOL, and OL for the immediate-threshold loss. This highlights the instability of generalized coefficients for the immediate-threshold loss. While a separate figure is not included, the results were similar under the original cumulative link loss.}
    \label{fig:my_label}
\end{figure}

\begin{figure}[!h]
    \centering
    \includegraphics[width=0.9\textwidth]{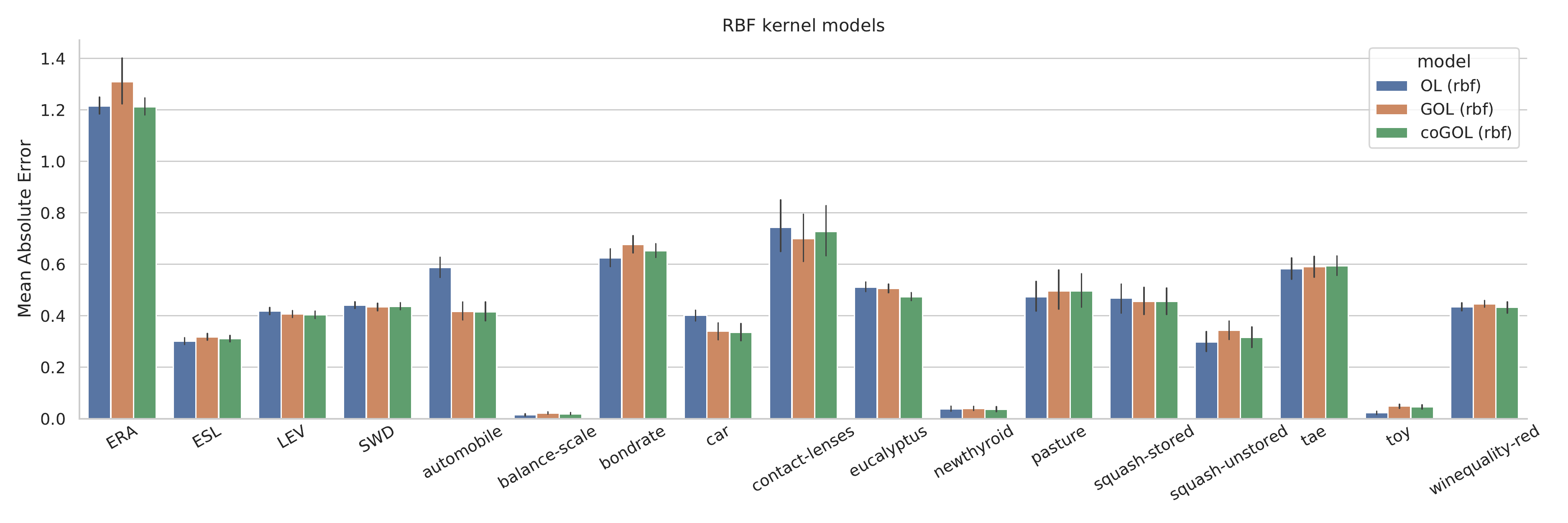}
    \caption{Comparison of kernelized coGOL, GOL, and OL with RBF kernel, with all-thresholds loss. Results are generally comparable to the all-thresholds linear models.}
    \label{fig:my_label}
\end{figure}

\newpage
\section{Base model for RF}

The components for the RF model were adapted from \cite{o2018over}. Each ResNet stack consists of:
\begin{enumerate}
    \item 1x1 convolution, input and output channels specified below.
    \item ResNet block 1:
        \begin{itemize}
            \item 1D convolution, kernel size 3, padding 1
            \item 1D batch norm, ReLU
            \item 1D convolution, kernel size 3, padding 1
            \item 1D batch norm
            \item Add the identity
        \end{itemize}
    \item ResNet block 2 (same)
    \item 1D max pooling, kernel size 2
\end{enumerate}
With the exception of the first stack, which has 2 input and 32 output channels, all subsequent stacks have 32 input and output channels.

The RF model has 6 such ResNet stacks, followed by 3 128-neuron FC layers with SELU activation and AlphaDropout with p=0.3.



\end{document}